\documentclass[letterpaper, 10 pt, conference]{ieeeconf}  

\IEEEoverridecommandlockouts                              

\usepackage{graphics} 
\usepackage{amsmath} 
\usepackage{amssymb}  
\usepackage{cite}
\usepackage{breqn}
\usepackage{stfloats}

\title{\LARGE \bf
A Contact-Safe Reinforcement Learning Framework for Contact-Rich Robot Manipulation
}

\author{Xiang Zhu, Shucheng Kang and Jianyu Chen*
\thanks{Xiang Zhu, Shucheng Kang and Jianyu Chen are with the institute for interdisciplinary information sciences, Tsinghua University, Beijing, China. Jianyu Chen is also with the Shanghai Qizhi Insitute, Shanghai, China. (*corresponding author, Email: {\tt\small jianyuchen@tsinghua.edu.cn})}%
}

\begin{document}

\maketitle
\thispagestyle{empty}
\pagestyle{empty}

\begin{abstract}
Reinforcement learning shows great potential to solve complex contact-rich robot manipulation tasks. However, the safety of using RL in the real world is a crucial problem, since unexpected dangerous collisions might happen when the RL policy is imperfect during training or in unseen scenarios. In this paper, we propose a contact-safe reinforcement learning framework for contact-rich robot manipulation, which maintains safety in both the task space and joint space. When the RL policy causes unexpected collisions between the robot arm and the environment, our framework is able to immediately detect the collision and ensure the contact force to be small. Furthermore, the end-effector is enforced to perform contact-rich tasks compliantly, while keeping robust to external disturbances. We train the RL policy in simulation and transfer it to the real robot. Real world experiments on robot wiping tasks show that our method is able to keep the contact force small both in task space and joint space even when the policy is under unseen scenario with unexpected collision, while rejecting the disturbances on the main task. 

\end{abstract}

\section{INTRODUCTION}

As the demand of automation in both industrial and daily applications increase rapidly, robots are required to accomplish more and more complex manipulation tasks. These tasks generally involve rich contacts with various unknown objects in unstructured environments, which makes the modeling and control of the robots quite challenging. 
Reinforcement Learning (RL), on the other hand, provides a model-free paradigm to solve complex tasks in unknown environments by self exploration and reinforcement. 
Several recent works have show great potential in applying RL to learn contact-rich robot manipulation policies,
such as grasping~\cite{kalashnikov2018scalable}\cite{levine2018learning}, insertion~\cite{beltran2020learning}\cite{spector2020deep} and assembling ~\cite{thomas2018learning}\cite{luo2019reinforcement}. 

\begin{figure}
\centering
\includegraphics[width=0.48\textwidth]{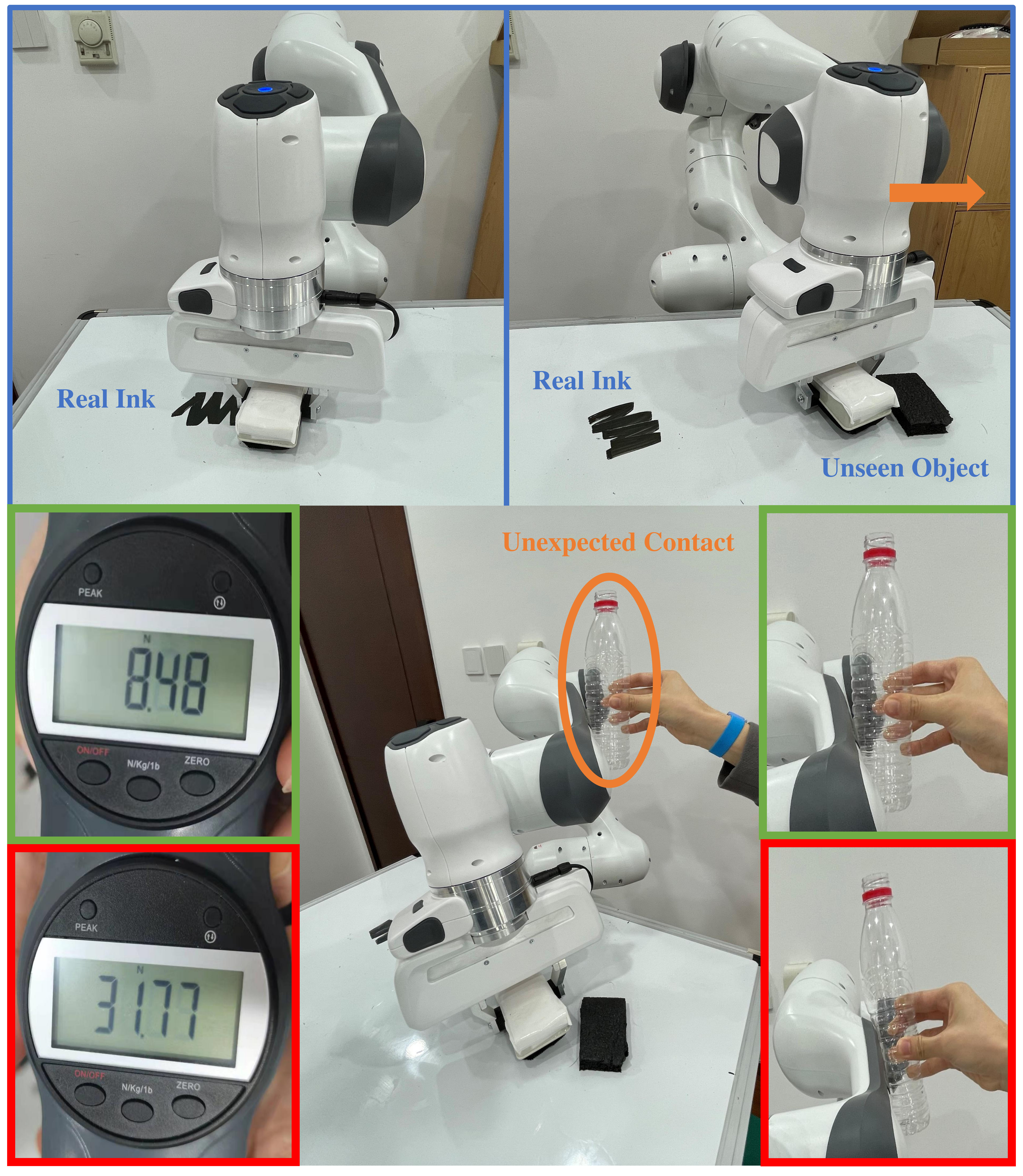}
\caption{Safety under unseen scenario: Our trained RL agent can finish the wiping task properly (shown in the top left figure). However, when an unseen object appears (shown in the top right figure), the RL policy fails and unexpectedly command the robot to move right. Such behavior might cause some unexpected contact. In this experiment, the robot collides with a plastic water bottle we grab. The green border pictures show that our method generate compliant contact behavior with small contact force. However the baseline, as shown in the red border figures, generate nearly 3 times the contact force of our method. As we can see, the baseline method significantly deforms the water bottle while our method does not deform it at all. The bottom left two figures show the contact force using a thrush meter and the bottom right two figures show the deformation of the water bottle.}
\label{fig:main fig}
\vspace{-3mm}
\end{figure}

Although quite promising, a major reason that prevents the wide use of RL in real world applications is its lack of safety. 
During the exploration procedure of RL, a common approach is to take some random actions~\cite{sutton2018reinforcement} or try to visit unexperienced states~\cite{burda2018exploration}, which, however, may lead to some unexpected dangers. Even if the training process of RL converges, the resulted policies are still prone to generate unexpected dangerous behaviors when met with unseen scenarios, and easily get interfered by external disturbances. For example, as shown in Fig. \ref{fig:main fig} (the two upper subfigures), an RL policy has been trained to wipe the ink on the whiteboard on the left. However, when there is an unseen object on the whiteboard, the policy is disturbed and the robot moves to the right. These unexpected behaviors of RL may result in unexpected collisions accompanying with large contact force between the robots and the environments (or even people), causing huge damages. Therefore, in this work we focus on such contact-safety when applying RL to contact-rich robot manipulation. 

According to where the dangerous contact might happen, the contact-safety can be categorized into task space safety and joint space safety. The task space safety requires the contact to be compliant between the robot end-effector and the targeted objects on which it is performing the tasks. 
Such contact compliance can be achieved with impedance control~\cite{hogan1985impedance}, 
which aims to control the dynamic relationship between the robot and the contact objects as a second order system consisting of a virtual mass, spring and damper. Variable impedance control (VIC)~\cite{abu2020variable} allows adjusting the stiffness according to demand, showing stronger potential for ensuring contact safety. Combining with RL,~\cite{martin2019variable} shows that using VIC in end-effector space as action space results in better sample efficiency, low energy consumption and more compliant end-effector contacts. Other related works include combining VIC and RL to perform high-precision robotic assembly tasks~\cite{luo2019reinforcement}, and combining model-based RL and VIC to enable safe collaboration between the robot's end-effector and human~\cite{roveda2020model}. However, all these works care about the task space safety problem solely, while ignoring the crucial joint space safety problem in RL.

Joint space safety focuses on the potential dangerous contacts happening at the robot arm, which can be common in human-robot interaction tasks or unstructured environments. 
Existing related works target on totally avoiding contact when using RL. Q. Yang et al.~\cite{yang2021null} ensure collision avoidance safety by only learning policies in the null space of higher prioritized constraints. S. Shaw et al.~\cite{shaw2021rmps} combine VIC and Riemannian motion policies (RMPs) which use geometric information of the arm and the surrounding environment to generate collision-free joint actions. ~\cite{zhao2021efficient}\cite{el2020towards} focus on the Human-Robot Collaboration scenario which also aims to find collision-free actions.
However, avoiding all collisions requires precise geometric information about the object, which is expensive and difficult to obtain. Furthermore, in cluttered and unknown environments, unexpected contacts might be inevitable. In some certain scenarios, we even need to create contact to better accomplish tasks~\cite{eppner2015exploitation}. In these cases, the joint space safety should be achieved not by avoiding collisions but by limiting the magnitude of the contact force. Nevertheless, to the best of our knowledge, no prior works have considered such joint space contact safety when using reinforcement learning.

In this paper, we propose a contact-safe reinforcement learning framework for contact-rich robot manipulation tasks, where both the task space safety and joint space safety are handled when using RL. Our framework is compatible with any RL algorithms, while in this work we use an image-based RL algorithm which learns a convolutional neural network (CNN) policy taking images as inputs and performing surface wiping tasks. The policy's output belongs to the Cartesian variable impedance control action space which helps generate compliant contact at the end-effector. A momentum observer is developed to estimate the contact point and force applied on the robot arm, purely from joint measurements. When collision is detected, a contact-aware controller will be activated, which performs a contact-aware null space projection that significantly reduces the contact force (as shown in Fig. \ref{fig:main fig}). Meanwhile, the disturbance compensation impedance term helps reject the disturbance to the main task. The RL policy is trained in simulation and transferred to the real robot. Real world experiment results show that compared to the baseline, our method produces significantly more compliant contact force during collision and less error during disturbances.

\section{Preliminary}
\subsection{Reinforcement Learning}
The goal of Reinforcement Learning is to find a reward maximizing policy $\pi$ that conditioned only on the current time step~\cite{sutton2018reinforcement}.
This problem can be modeled as a Markov decision process (MDP) which is represented by a tuple $(\mathcal{S}, \mathcal{A}, \mathcal{P}, \mathcal{R}, \gamma, \mu)$, 
where $\mathcal{S}$, $\mathcal{A}$ are the state and action spaces respectively,
$\mathcal{P}$ represents the transition model that maps the state and action into a probability distribution of the next state $P(s'|s,a)$, $\mathcal{R}$ represents the reward function $r(s,a)$, $\gamma$ is the discount factor for the reward function and $\mu$ represents the initial state distribution. We also define the discounted future return $R\left( \tau \right) =\sum\nolimits_{i=t}^T{\gamma ^{i-t}r\left( s_i,a_i \right)}$, where $\tau =\left( s_t,a_t\cdots s_T,a_T \right)$. Assume our policy $\pi$ is parameterized by $\theta$, then the RL algorithm optimizes $\theta$ by maximizing the expected future return:
\begin{equation}\label{eq: 1}
\theta ^*=\underset{\theta}{\text{arg max}} \ \mathbb{E} _{\tau \sim \pi \left( \theta \right)}R\left( \tau \right) 
\end{equation}

\subsection{Operational Space Formulation and Varible Impedance Control}

Operational space formulation~\cite{khatib1987unified} is widely used in torque controlled robots. 
The equation of motion of an open chain robot manipulator can be written as:
\begin{equation}
    M(q)\ddot q + C(q, \dot q) + g(q) = \tau_m - \tau_{ext}
\end{equation}
where $q \in \mathbb{R}^n$ represents the generalized coordinates, $M(q) \in \mathbb{R}^{n\times n}$ is the positive-definite and symmetric mass matrix, $C(q,\dot q) \in \mathbb{R}^n$ is the centrifugal and Coriolis term, $g(q)\in \mathbb{R}^n$ is the gravity vector, $\tau_m \in \mathbb{R}^n$ and $\tau_{ext} \in \mathbb{R}^n$ on the right-hand side is the applied torque and the external joint torque.

Considering an manipulator with task coordinates $t$, we can write the mapping from generalized coordinates to task coordinates as $\dot x = J_t(q) \dot q$, where $J_t$ is the task Jacobian and is assumed be full row-rank. We can write the task mass matrix as $\Lambda_t(q) = \left(J_t(q) M(q)^{-1}  J^T_t(q) \right)^{-1}$. 
The dynamically consistent inverse of the task Jacobian can be written as:
\begin{equation}
    J^{\#}_t(q) = M(q)^{-1} J^T_t(q)  \Lambda_t(q)
\end{equation}

Multiplying the equation of motion in the generalized coordinates by the transpose of the dynamically consistent inverse of the task Jacobian, we can get the equation of motion written in the operational space:
\begin{equation}
    \Lambda_t(q) \ddot x + \mu_t(q, \dot q) + p_t(q) = F_t - \left( J^{\#}_t(q) \right) ^T \tau_{ext}
    \label{eq:operational space dynamic}
\end{equation}
where $\mu_t$ is the task space Centrifugal and Coriolis force and $p_t$ is the gravity vector project into the task space. The command joint torque is give by $\tau_m = J^T_t(q) F_t$.

Following~\cite{martin2019variable} who claim that using variable impedance control in end-effector space have intrinsic advantages, such as sample efficiency and ensures task space safety, we use variable impedance control in task space to calculate the task control force $F_t$.
\begin{equation}
\begin{split}
F_t = &\Lambda_t(q) \left( K_1(x_{d}-x)-D_1 \dot x \right) + \mu_t(q, \dot q) + p_t(q) \\
     & + \left( J^{\#}_t(q)\right)^T \tau_{ext}
\end{split}
\end{equation}
$K_1$ and $D_1$ are the diagonal stiffness and damping matrix generated by the RL policy. The equilibrium point or the desired pose $x_{d}$ is also given by the policy.

\subsection{Null Space Projection}
The redundant degrees of freedom (DoF) allow the robot to achieve more dexterous motions. In robotics, the most commonly used method to take advantage of kinematic redundancy is the null space projection method~\cite{nakamura1987task}\cite{dietrich2015overview}.

Considering a robot with a prior task and a second task, the goal of null space projection is to execute the second task as good as possible without disturbing the prior task.
The Jacobian of the prior task is defined as $J_1(q)$. We can use the successive null space projection~\cite{antonelli2009stability} to project the second task torque $\tau_2$ into the null space of the prior task,
\begin{equation}
    \tau_2^p = N^T_2(q)\tau_2
\end{equation}
where $\tau_2^p$ is the projected torque that does not disturb the prior task.
The successive null space projector $N_2(q)$ is defined as:
\begin{equation}
    N_2(q) = I - J^{\#}_1(q) J_1(q)
\end{equation}
where $J^{\#}_1(q)$ is the generalized inverse of $J_1(q)$ which satisfies $J_1(q)J^{\#}_1(q) = I$. We can compute $J^{\#}_1(q)$ by
\begin{equation*}
    J^{\#}_1(q) = W^{-1}J^T_1(q)\left(J_1(q)W^{-1}J^T_1(q)\right)^{-1}
\end{equation*}
where the common choice of $W$ is the identity matrix $I$ or the mass matrix $M(q)$.

\section{Methods}

\begin{figure*}[htbp]
\centering
\includegraphics[width=1.0\textwidth]{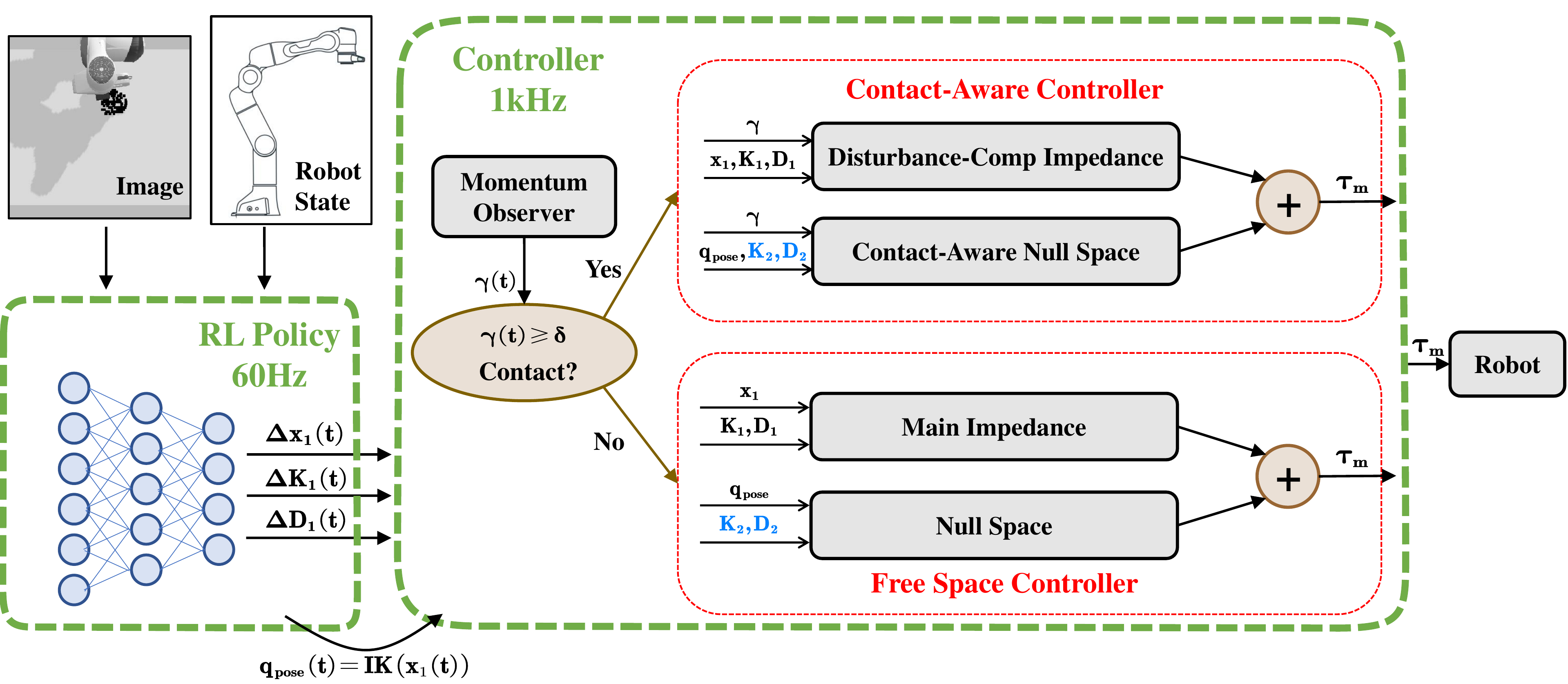}
\caption{Framework Overview: Our method consists of a low-rate RL policy and a high-rate controller. The RL policy takes the image and the robot eff-effector pose as input, and outputs task space information. We use inverse kinematics to define the posture task. The controller consists of a contact-aware controller and a free space controller while using momentum observer as the contact detector. The controller switches to the contact-aware controller under contact, and switches to the other on the contrary. The final command torque will be sent to the robot.}
\vspace{-3mm}
\label{fig:framework}       
\end{figure*}

The framework overview of our proposed method is shown in Fig. \ref{fig:framework}. Our framework is mainly composed of two parts, the RL policy and the controller. The RL policy takes both the camera observation image and the robot state as inputs, and outputs the increment of end-effector pose, as well as the stiffness and damping matrices of the Cartesian VIC. The controller takes the output of the RL policy as the input and outputs the robot joint torques. A modified momentum observer from~\cite{de2005sensorless}\cite{haddadin2017robot} is used as the contact estimator to detect unexpected collisions on the robot arm. When no collision is detected, the controller will execute the main VIC task which maintains task space safety, while performing a posture task in the null space to regulate the joint behavior. Once collision is detected, a contact-aware controller will be activated, which will perform contact-aware null space projection to keep the external contact force small, while executing a disturbance compensation impedance task to reject the effects of external disturbances on the main task. The RL policy and the controller run at different frequencies due to the computation resource limitation. The policy can be obtained with any RL algorithm while in this work we use proximal policy optimization (PPO)~\cite{schulman2017proximal} with variational autoencoder (VAE)~\cite{higgins2016beta} pretraining. We now describe the details of our proposed method.


\subsection{Image-Based RL Policy}
In this paper, we consider a generic setting of image-based RL policy, where both the observation image and the robot state are used as the policy inputs. To improve the sampling efficiency, we firstly use a variational auto encoder (VAE)~\cite{higgins2016beta} to extract the latent embedding for the image in an unsupervised way. Let us denote the image as $o$, then the latent embedding $s_{img}$ obtained from the VAE's encoder $f_\phi$ is defined as:
\begin{equation}
    s_{img} = f_\phi(o)
\end{equation}
The final state $s$ is obtained from concatenating the end-effector's 6-DoF pose $s_{pose}$ with $s_{img}$. Putting the state $s$ into PPO's policy network $\pi_\psi$, we get the action $a$:
\begin{equation}
a=\pi_\psi \left( s \right) =\pi \left( \left[ \begin{array}{c}
	s_{pose}\\
	s_{img}\\
\end{array} \right] \right) 
\end{equation}
where we use variable impedance control in Cartesian space as our action space which consists of the increment of the diagonal stiffness matrix $K_1$, the damping matrix $D_1$ and the equilibrium point $x_d$. Details of the training process of our RL policy will be introduced in section III.F.

\subsection{Contact Estimator}
To estimate whether a contact is happened on the robot arm, we use the momentum observer~\cite{de2005sensorless}\cite{haddadin2017robot}, which is a stable, linear, decoupled, first-order estimation method of the external joint torque $\tau_{ext}$. Define the generalized momentum $p$ of the robot as
\begin{equation}
    p = M(q) \dot q
\end{equation}

\cite{de2005sensorless}\cite{haddadin2017robot} suggest that the $\tau_{ext}$ can be approximated using the residual vector $\gamma \in \mathbb{R}^n$:
\begin{equation}
    \gamma = K_I\left( p(t) - \int^t_0 (\tau_m - \hat \beta(q, \dot q) + \gamma) ds - p(0) \right)
\label{origin obs}
\end{equation}
where $\hat \beta$ is the estimate of $\beta(q, \dot q)=g(q)+C(q, \dot q) - \dot M(q) \dot q$ and $K_I > 0$ is the diagonal gain matrix.
Under ideal conditions that mass matrix $M(q)$ and $\beta(q, \dot q)$ is well estimated, the dynamics of the residual vector and the external joint torque is given by
\begin{equation}
    \dot \gamma = K_I(\tau_{ext} - \gamma)
\end{equation}
If $K_I \to \infty$, we have $\gamma \approx \tau_{ext}$.

Using the above method we can estimate the total external joint torque, which consists of torques induced by both the unexpected contact force happened at the robot arm and the expected contact force happened at the end-effector:
\begin{equation}
    \tau_{ext} = \tau_c + \tau_{eff} = J^T_c(q) F_c + J^T_t(q) F_{eff}
\end{equation}
where $J_c(q)$ is the Jacobian at the contact point, $F_c$ is the contact force at joint level, $\tau_c$ is the corresponding contact joint torque and $F_{eff}$ is the conact force at the end-effector. Usually $F_{eff}$ can be directly measured by the force torque sensor. If we want to estimate the contact force applied on the arm only, we need to cancel out the end-effector contact force in the residual vector:
\begin{equation}
    \begin{aligned}
    \gamma = & K_I \left[ p(t) - \int^t_0 \left(\tau_m + J^T_{t}(q) F_{eff} - \hat \beta(q, \dot q) + \gamma \right)ds \right.\\
     & \left.- p(0)\right]
    \end{aligned}
\label{our obs}
\end{equation}





In the free space, the residual vector should ideally be zero. But in practice, it's hard to get a precise estimation, so we assume the contact condition is triggered when the residual is greater than a predefined threshold $\delta$. 
The link in contact corresponds to the coefficient of $\gamma$ that is greater than $\delta$ and is the furthest away from the robot base.


\subsection{Free Space Controller}
Using variable impedance control in the end-effector space has many intrinsic benefits, but this appealing control scheme would generate unpredictable motion in the free space. As suggested in~\cite{luo2019reinforcement}, we find the same problem that the arm might unexpectedly continue to accelerate. We name this problem as the joint acceleration problem when no joint task exist. The reason is that the joint control torque $\tau_{ext}$ is a projection of the task control force $F_t$, and there are no control targets on the robot joints. To alleviate this problem, we add an additional posture task to the robot joints. 

In order to prevent the posture task from affecting the execution of the prior task, we use null space projections to execute the posture task as good as possible without disturbing or interfering the prior task. The posture task goal is to stay in a reference configuration $q_{pose}$:
\begin{equation}
    \Gamma_{pose} = K_2(q_{pose}-q) - D_2\dot q
\end{equation}
where $K_2$ and $D_2$ are the diagonal joint stiffness and damping matrix. The dynamic consistent null space projection matrix is:
\begin{equation}
    N_t(q) = I - J^{\#}_t(q) J_t(q)
\end{equation}

This projector can project the posture force into the null space of the prior task. Note here the $\tau_{ext}$ term in Equation (\ref{eq:operational space dynamic}) is zero. The free space controller is defined as:
\begin{equation}
    \tau_m = J^T_t(q) F_t + N^T_t(q) \Gamma_{pose}
    \label{Freespace control}
\end{equation}

\subsection{Contact-Aware Controller}
Using the free space controller with a well-trained RL policy will maintain the task space safety. However, the robot sometimes will encounter unexpected contact at its arm, during human interaction or working in unstructured environment. In order to keep the joint space safety, we design a contact-aware controller inspired by~\cite{jorda2019contact}. We use a disturbance compensation impedance to avoid the unexpected contact from disturbing the prior task, and a contact-aware null space projection to keep the contact force small between the robot arm and the environment.

\subsubsection{Disturbance Compensation Impedance}
In free space, the contact joint torque $\tau_c$ is zero. But under joint contact situation, we need to consider the effect of $\tau_c$. The task control force can be written as:
\begin{equation}
\begin{aligned}
    F_{t-c} = &\Lambda_t(q) \left( K_1(x_{d}-x)-D_1 \dot x \right) + \mu_t(q, \dot q) + p_t(q) \\
     & - \left( J^{\#}_t(q) \right) ^T \gamma
\end{aligned}
\end{equation}

In order to avoid disturbing the prior task defined by the Cartesian variable impedance control, we need to cancel out the joint torque induced by the expected contact force at the end-effector.
Therefore, we use Equation~(\ref{our obs}) to calculate $\gamma$, which only estimate the contact joint torque $\tau_c$.

\subsubsection{Contact-Aware Null Space Projection}
When in contact with the environment, the robot is under additional constraints which makes the robot lose degrees of freedom in the motion space. To prevent the robot from violating the constraint, we need to remove this degree of freedom from the posture task force. At the contact point, we have:
\begin{equation}
    \tau_c = J^T_c(q) F_c
\end{equation}
where $J_c(q)$ is the Jacobian at the contact point, $F_c$ is the contact force and $\tau_c$ is the corresponding contact joint torque. 
The direction of the contact can be written as
\begin{equation}
    u_c:= \frac{F_c}{||F_c||}\quad\;\; J_{u_c} = u_c^T J_c(q)
\end{equation}
The reduced contact Jacobian $J_{u_c}$ maps the joint velocity $\dot q$ to the velocity along the direction of the contact $u_c$.
If we project the posture task into the intersection of the null space of the task Jacobian and that of the reduced contact Jacobian~(namely the task-consistent null space of the reduced contact Jacobian), the posture task will not violate the contact constraint as well as the prior task, which results in compliant behavior.

Practically, the task-consistent null space of the reduced contact Jacobian can be calculated as follows~\cite{jorda2019contact}:
\begin{equation}
    N_{c|t} = N_t \left( I - (\gamma^T N_t)^{\#}_{M^{-1}}(\gamma^T N_t) \right)
\end{equation}
where $(\gamma^T N_t)^{\#}_{M^{-1}}$ is the generalized inverse of $\gamma^T N_t$ weighted by the inverse of the mass matrix.

Now the control law becomes:
\begin{equation}
    \tau_m = J^T_t(q) F_{t-c} + N_{c|t}^T \Gamma_{pose}
\end{equation}
\label{contact control}
The total control law of our controller part is:
\begin{itemize}
    \item No contact~(Free Space Controller): 
    \begin{equation*}
        \tau_m = J^T_t(q) F_t + N^T_t(q) \Gamma_{pose}
    \end{equation*}
    \item With contact~(Contact-Aware Controller): 
    \begin{equation*}
        \tau_m = J^T_t(q) F_{t-c} + N_{c|t}^T \Gamma_{pose}
    \end{equation*}
\end{itemize}

\subsection{Coordinates between RL Policy and controller}

As shown in Fig.\ref{fig:framework}, our framework is hierarchical, where the upper level is the RL policy and the lower level is the controller. The RL policy generates the increment of the equilibrium point, the task space diagonal stiffness and damping matrix. After obtaining $x_{d}$ from the policy, we can perform inverse kinematics (IK) to define the posture task:
\begin{equation}
    q_{pose} = \text{IK}(x_{d})
\end{equation}

The high-level RL policy runs at a lower rate (e.g. 60Hz), while the low-level torque controller runs at a higher rate (e.g. 1000Hz). We observe that the controller has a jittering phenomenon under some disturbance contact caused by: (1) the oscillation of the residual vector and (2) the switching between the two sub-controllers. We add a low-pass filter and set the gain $K_I$ of the residual vector small to mitigate the first problem. We alleviate the second problem by enforcing that the controller will not be switched until the execution time of the contact-aware controller is greater than the threshold $t_{c}$.

\subsection{Training of RL Policy}
We train our policy in the simulation with no unexpected contact on the robot arm. We first pretrain a VAE model from randomly sampled possible observations with the loss function:
\begin{equation}
    \begin{aligned}
        \mathcal{L}_{\text{VAE}}(\theta, \phi ; o, s_{img}, &\beta)= \mathbb{E}_{q_{\phi}(s_{img}|o)}\left[\log p_{\theta}(o|s_{img})\right] \\ 
        & -\beta D_{KL}\left(q_{\phi}(s_{img}|o) \| p(s_{img})\right)
    \end{aligned}
\end{equation}
where $D_{KL}(\cdot \| \cdot)$ is the KL Divergence between two distributions, $\theta$ is the parameters of the decoder network, the the encoder $f_\phi(\cdot)$ is the mean of the variational distribution $q_{\phi}(s_{img}|o)$ and $\beta$ is a  hyperparameter.

The RL policy is trained with the free space controller as described in section III.C using PPO~\cite{schulman2017proximal}. The training objectives is to maximize:
\begin{equation}
    \mathcal{L}(\psi)=\mathbb{E}_{t}\left[\min \left(r_{t}(\psi) A_{t}, \operatorname{clip}\left(r_{t}(\psi), 1-\epsilon, 1+\epsilon\right) A_{t}\right)\right]
\end{equation}
where $A_t$ is the advantage function, $\epsilon$ is a hyperparameter, $\psi$ is the parameter of the policy network. $r_{t}(\psi)$ is defined as:
\begin{equation}
    r_{t}(\psi)=\frac{\pi_{\psi}\left(a_{t} | s_{t}\right)}{\pi_{\psi_{\mathrm{old}}}\left(a_{t} | s_{t}\right)}
\end{equation}
where $\psi_{\mathrm{old}}$ is the vector of policy parameters before the update.

\vspace{2pt}
\section{Experiments}
Among multiple contact-rich tasks, we choose surface wiping as our experimental showcase. Our agent's goal is to wipe the ink drawn on the whiteboard. While wiping the surface, we intentionally cause unexpected contacts on the robot arm, and compare both task space safety and joint space safety with the baseline method. 

\subsection{Implementation Details}
We implement our method on a real Franka Panda 7 DoF robot arm with a two-finger gripper. We use AMD 5800H and NVIDIA RTX3070 as run-time CPU and GPU with an Intel RealSense D455 camera. We train our agent only in simulation, then deploy the learned policy in real environment with zero-shot. 
We choose pybullet~\cite{coumans2021} as our simulator. 

For each training episode, ink is randomly generated on the simulated whiteboard. The observation space incorporates both end-effector's 6-DoF pose and a 128$\times$128$\times$1 image. 
In order to close the sim2real gap of the image, we do segmentation both on real and simulation, where pixels inside the ink are set to zero while others are set to 255. For image feature extraction, we pretrain a VAE model, which encodes an image to a 128-dimension vector for the downstream RL tasks.
Our agent gets a reward proportional to the inks wiped off, and an additional reward when all inks are wiped. 
To encourage exploration at the beginning, a small reward is given if the agent keeps contacting the table and the contact force is less than $40N$. Moreover, to guide the end-effector toward the ink, we provide a small reward increasing as the distance between the end-effector's position and the ink decreases. When the robot arm collides with the table, we give a large penalty and the episode terminates.

When transferred to the real world, the RL policy runs at 60Hz and the controller runs at 1000Hz. 
We use~\cite{beeson2015trac} as our inverse kinematics solver. The momentum observer threshold $\delta$ is $1.5 Nm$ and diagonal gain matrix $K_I$ is set as $1.5I$ where $I$ is the identity matrix. For posture task, we set the stiffness $K_2$ for each joint as $\left[40, 40, 40, 40, 30, 20, 10\right]$ and set the damping $D_2$ for each joint as $\left[12, 12, 12, 12, 10, 8, 6\right]$. The controller switch threshold $t_{c}$ is set as $50ms$.

Our baseline follows the VICES framework~\cite{martin2019variable}, except that a posture task is added for avoiding the accelerate problem, and all the parameters are set the same with our method for a fair comparison. In general, VICES considers task space safety but does not consider joint space safety.

\vspace{-10pt}
\begin{figure}[hb]
\centering
\includegraphics[width=0.485\textwidth]{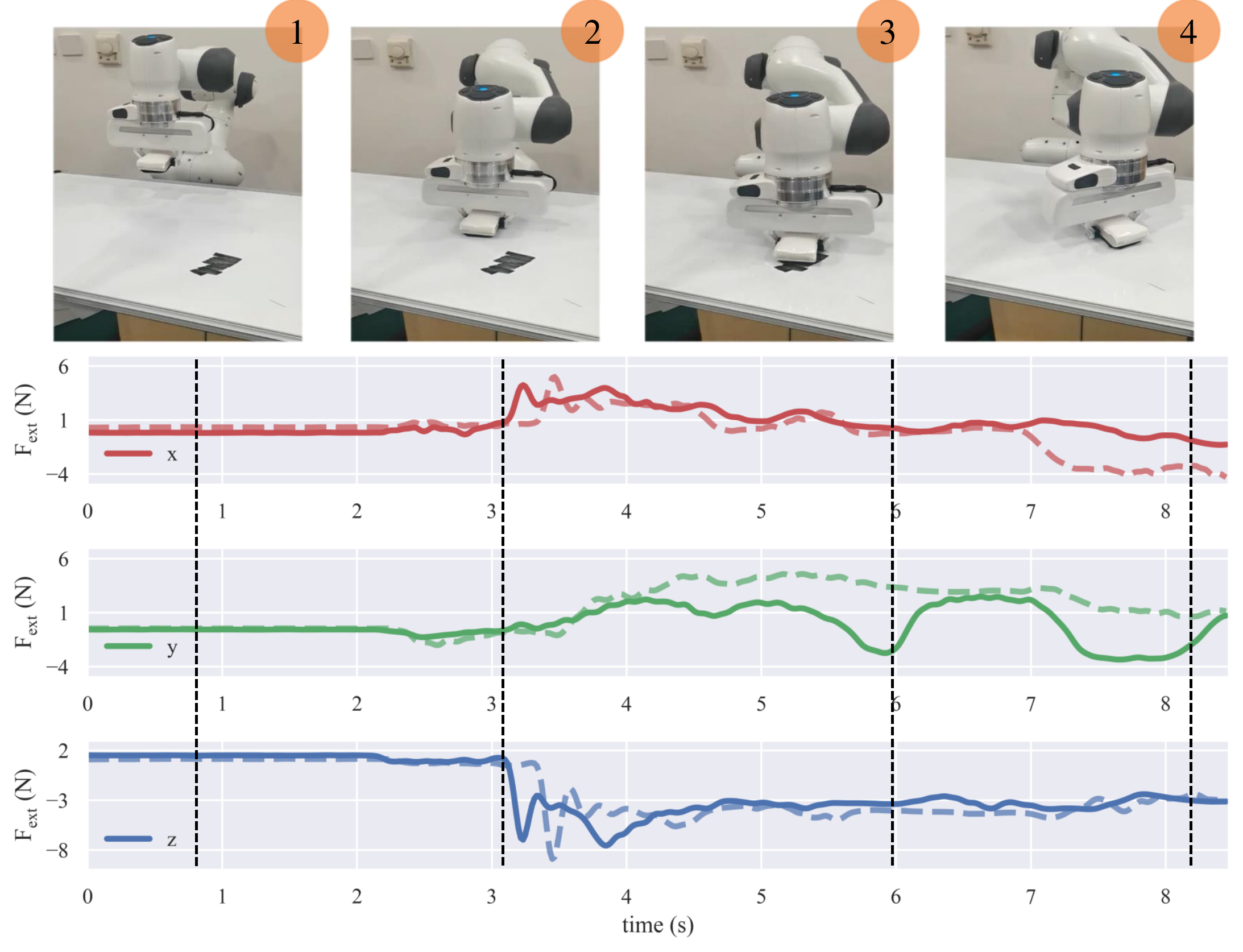}
\vspace{-25pt}
\caption{Task Space Safety: The bottom three figures plot the 3-axis end-effector force when the robot wipes the whiteboard. The safe task contact is defined as the contact force under $40N$. The maximum force of our methods is about $5N$ which is far less than the safe contact threshold, while performing the wiping task successfully. The black dashed lines mark the time when the situation shown in the top images occurred and other colored dashed line plots the result of the baseline.}
\vspace{-15pt}
\label{fig:exp1}       
\end{figure}

\subsection{Task Space Safety}

We firstly show the task space safety of our proposed method, which is defined by how large the end-effector contact force is when performing the contact-rich manipulation task. We manually draw a set of inks on the table and let the robot wipe the inks actuated by the trained RL policy and the controller. The result is shown in Fig. \ref{fig:exp1}. The maximum force of the Franka Panda robot arm can apply is $40N$. In our setting, same as VICES, the unsafe situation is defined as if the task contact force at the end-effector is greater than $40N$. 
We can see that the robot can finish the wiping task successfully while keeping a safe contact force. Note both our proposed method and the baseline maintains compliant contact since the baseline also considers task space safety. 
Our method shows similar performance with the baseline since their only difference is the use of posture task when focusing on task space safety. This result shows that the posture task does not affect the main task.

\begin{figure}[ht]
\centering
\includegraphics[width=0.48\textwidth]{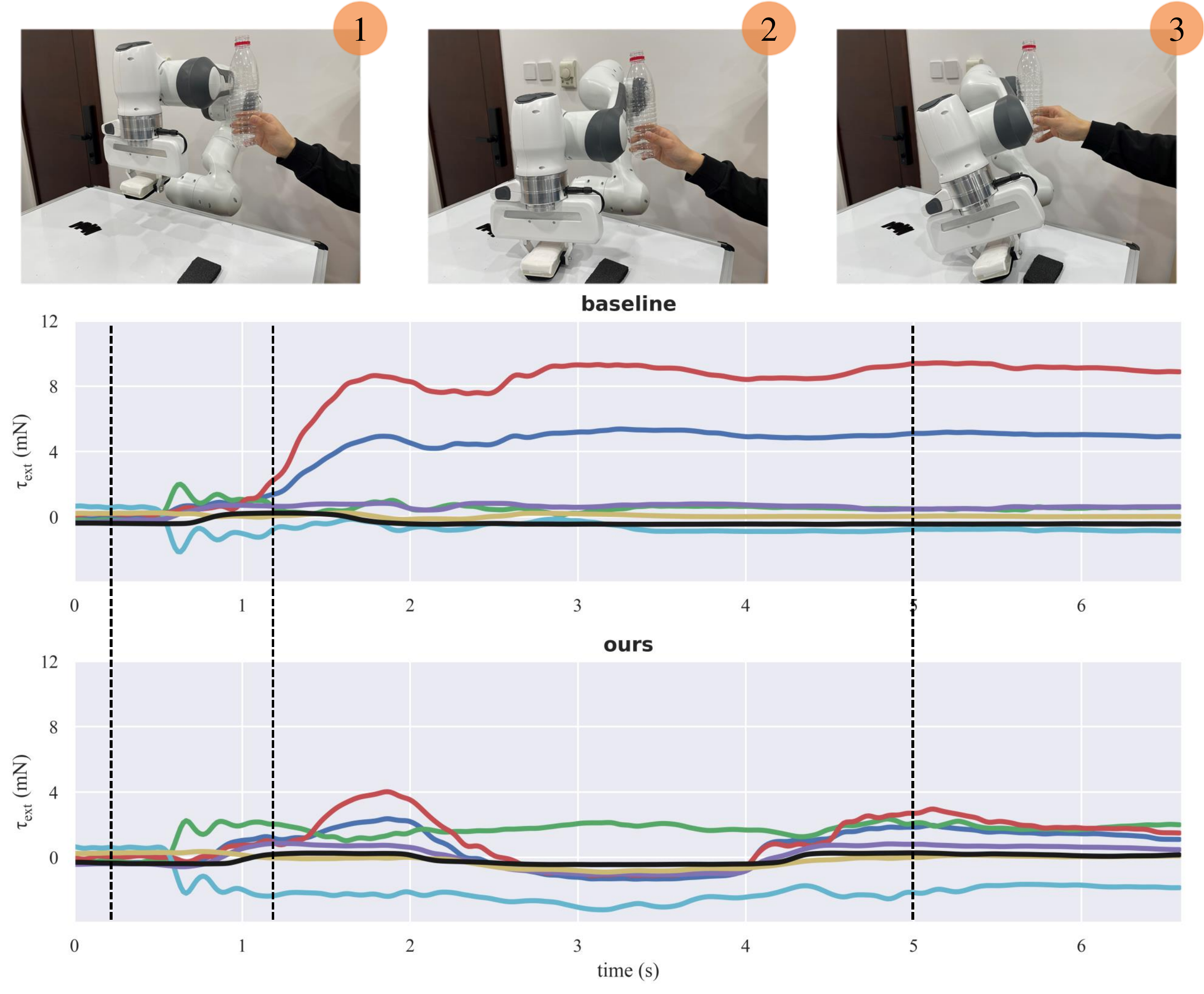}
\vspace{-25pt}
\caption{Joint Space Safety: The bottom two figures plot the estimated external joint contact force induced by unexpected contact at the robot arm.}
\vspace{-5mm}
\label{fig:exp2}       
\end{figure}

\subsection{Joint Space Safety}

To illustrate that our method can maintain the joint space safety with an RL policy, we hold a plastic water bottle at the position where the robot will pass through as the robot see the unseen black styrofoam, as shown in Fig. \ref{fig:main fig}. In our experiment, the robot will successfully wipe the ink if no black styrofoam was put in place. However, if the black styrofoam was placed, the robot will go right and hit the plastic water bottle we hold. The black styrofoam thus acts as a disturbance for the RL agent. 

The bottom figure in Fig. \ref{fig:main fig} intuitively shows the contact force. The green border is ours and the red one is the baseline. Our method does not deform the plastic water bottle at all while the baseline deforms it significantly.
We also use a thrust meter to measure the actual contact force quantitatively at the contact point, holding the thrust meter at the same position as the water battle.
The contact force of our method shown in the thrust meter is 3 times less than the baseline. Fig. \ref{fig:exp2} shows that the estimated external joint contact force in our methods is less than the baseline: the baseline reaches $10Nm$ but our method is less than $5Nm$.

We can see that our method generates far less contact force and shows a much more compliant behavior. This compliant behavior can provide better safety guarantees, which is crucial for RL agents in the real world, especially when the agent visits an unexperienced state, or is in the exploration phase where random actions may lead to dangers. Our compliant controller protects both the robot and the surrounding environment (including humans).

\subsection{Task Disturbance Rejection}
\begin{figure}[htbp]
\centering
\includegraphics[width=0.47\textwidth]{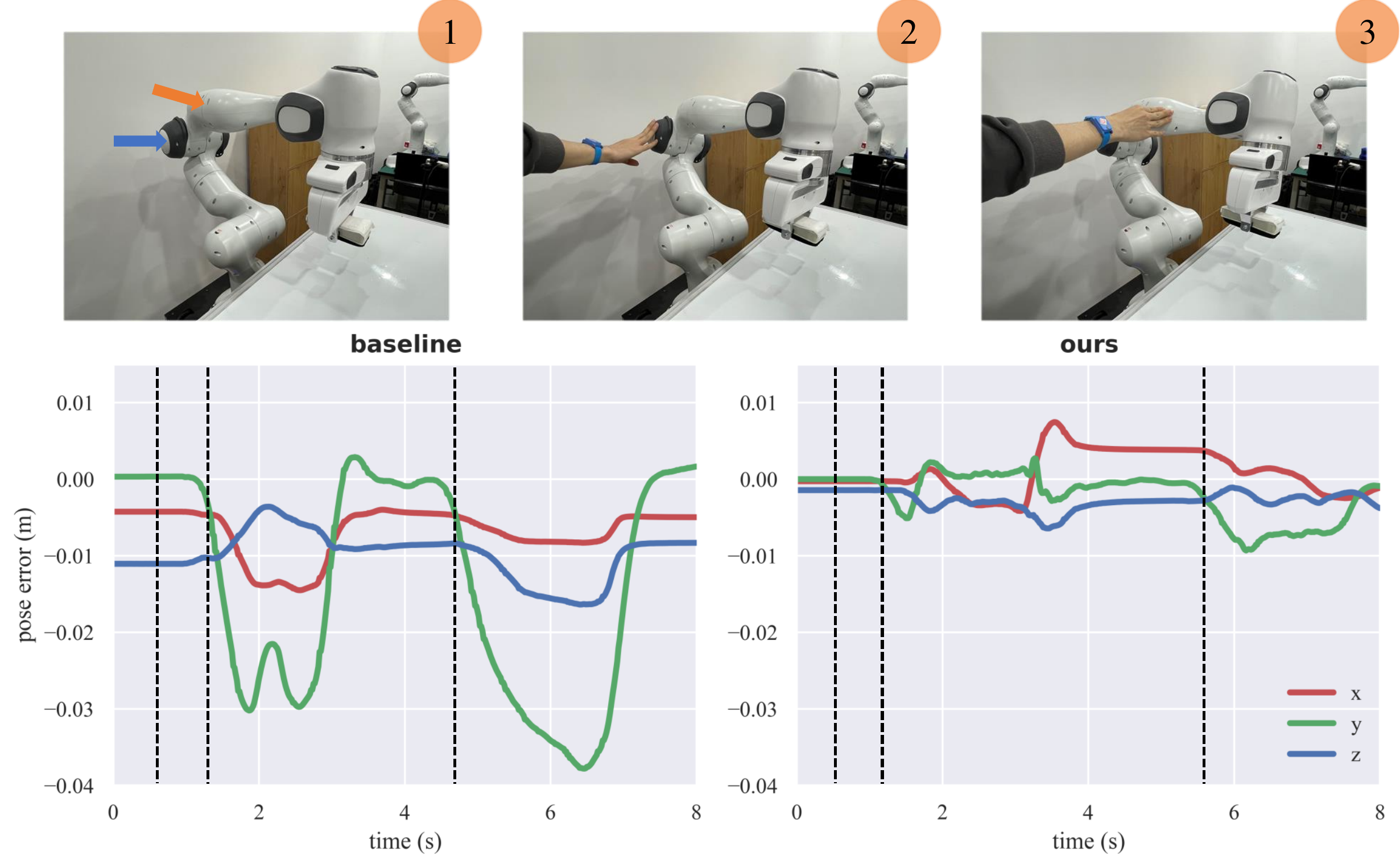}
\vspace{-15pt}
\caption{Task Disturbance Rejection: The blue and orange arrows show the disturbance force applied by human. The blue one applied first while the orange one applied next. The bottom right figure shows the end-effector position error of our method under the unexpected contact. The bottom left one shows the result of the baseline. }
\vspace{-5mm}
\label{fig:exp4}       
\end{figure}

When undergoing unexpected contacts, it is desirable if the robot can perform a compliant behavior while rejecting disturbances and keep the prior task as precise as possible. We show that our method has better accuracy on the prior task while maintaining compliance. In our experiments, we randomly disturb the robot by pushing it when it is working, shown as the arrows in Fig. \ref{fig:exp4}. We can see that the baseline method is much more easier to get affected, reaching a $4cm$ maximum error in the $y$ direction. However, the error of our method is less than $1cm$ in all directions. This shows that our method can reject the impact disturbances on the prior task due to unexpected contacts. 

\vspace{-3mm}

\section{CONCLUSION}
This work focuses on using reinforcement learning in contact-rich manipulation tasks while keeping both task space and joint space contact safety, which is important due to the potential risks in both the exploration and deployment procedures of RL. We proposed a hierarchical framework with an RL policy to generate the cartesian VIC target, and a controller to perform this task while keeping compliant for both the end-effector task contact and the robot arm unexpected contact. We trained our policy in the simulation and deployed it on a real Franka Panda 7 DoF robot. The whiteboard wiping experiments show that our method is able to perform compliant behaviors under unexpected contacts caused by unexpected RL policy outputs. Results also suggest that the proposed framework maintains safe contact force when performing the wiping task, while rejecting the disturbances caused by unexpected joint contacts. Although focusing on the wiping task using a specific RL algorithm in this work, our framework is generic and can be applied to other tasks with different RL algorithms.

\bibliographystyle{ieeetr}
\bibliography{cite}

\end{document}